\documentclass[10pt,twocolumn,letterpaper]{article}

\usepackage{cvpr}              

\usepackage{multicol,multirow}
\usepackage{xcolor}
\usepackage{booktabs}

\usepackage{colortbl}
\usepackage{floatflt}
\usepackage{float}
\usepackage{pifont}
\usepackage{makecell}
\usepackage{caption}
\usepackage{capt-of}
\definecolor{midgray}{gray}{0.5}
\newcommand{\boldres}[1]{{\cellcolor{gray!20}\textbf{#1}}}
\newcommand{\secondres}[1]{{\cellcolor{gray!10}\underline{#1}}}
\usepackage[most]{tcolorbox}
\usepackage{amssymb, pifont}

\usepackage{amsmath,amsfonts,bm}

\def\eqref#1{equation~\ref{#1}}

\def\1{\bm{1}}

\DeclareMathAlphabet{\mathsfit}{\encodingdefault}{\sfdefault}{m}{sl}
\SetMathAlphabet{\mathsfit}{bold}{\encodingdefault}{\sfdefault}{bx}{n}

\definecolor{cvprblue}{rgb}{0.21,0.49,0.74}
\usepackage[pagebackref,breaklinks,colorlinks,allcolors=cvprblue]{hyperref}

\def\paperID{5820} 
\def\confName{CVPR}
\def\confYear{2026}

\title{\textbf{\texttt{OmniFM}}: Toward Modality-Robust and Task-Agnostic \\ Federated Learning for Heterogeneous Medical Imaging}

\author{Meilin Liu, Jiaying Wang, Jing Shan
\\
School of Software, Shenyang University of Technology \\
{\tt\small lmeilin153@gmail.com, \{jiaying, shanjing\}@sut.edu.cn}
}

\begin{document}
\twocolumn[{
\renewcommand\twocolumn[1][]{#1}
\maketitle
\begin{center}
    \centering
    \includegraphics[width=.92\textwidth]{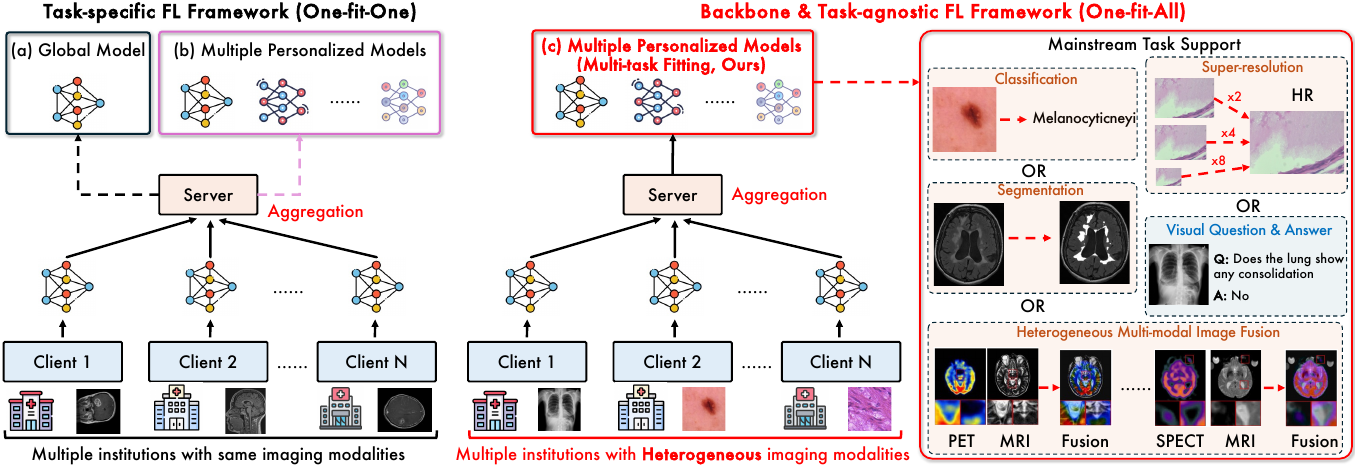}
    \captionof{figure}{Conventional federated learning for medical image is task-specific and assumes homogeneous imaging modalities. Our \textbf{\texttt{OmniFM}} offers a single reusable pipeline that supports diverse backbones and remains robust under heterogeneous modalities for multiple tasks.}
    \label{fig:intro_fig}
\end{center}
}]

\begin{abstract}
Federated learning (FL) has become a promising paradigm for collaborative medical image analysis, yet existing frameworks remain tightly coupled to task-specific backbones and are fragile under heterogeneous imaging modalities. Such constraints hinder real-world deployment, where institutions vary widely in modality distributions  and must support diverse downstream tasks. To address this limitation, we propose \textbf{\texttt{OmniFM}}, a modality- and task-agnostic FL framework that unifies training across classification, segmentation, super-resolution, visual question answering, and multimodal fusion without re-engineering the optimization pipeline. \textbf{\texttt{OmniFM}} builds on a key frequency-domain insight: low-frequency spectral components exhibit strong cross-modality consistency and encode modality-invariant anatomical structures. Accordingly, \textbf{\texttt{OmniFM}} integrates (i) Global Spectral Knowledge Retrieval to inject global frequency priors, (ii) Embedding-wise Cross-Attention Fusion to align representations, and (iii) Prefix–Suffix Spectral Prompting to jointly condition global and personalized cues, together regularized by a Spectral-Proximal Alignment objective that stabilizes aggregation. Experiments on real-world datasets show that \textbf{\texttt{OmniFM}} consistently surpasses state-of-the-art FL baselines across intra- and cross-modality heterogeneity, achieving superior results under both fine-tuning and training-from-scratch setups.
\end{abstract}

\section{Introduction}
Medical data are typically governed by strict privacy regulations (e.g., GDPR~\cite{gdpr2016general}, HIPAA~\cite{cohen2018hipaa}), preventing centralized aggregation across institutions. Unlike natural image corpora that can be scraped at scale, medical data must remain siloed, making it difficult to train high-capacity deep models in a centralized manner. Federated learning (FL)~\cite{mcmahan2017communication} has emerged as a promising paradigm that enables multi-institutional training while keeping data on-premise. Recent studies have demonstrated the feasibility of FL in mainstream medical imaging tasks such as classification~\cite{chen2025restyled}  segmentation~\cite{jiang2023fair}, and visual question answer offering a trustworthy collaboration protocol across organizations.

Despite these advances, existing FL frameworks remain largely task-specific and backbone-coupled, as shown in the left of \textbf{Fig.~\ref{fig:intro_fig}}. A federation configured for classification often assumes a CNN backbone and requires dedicated aggregation~\cite{li2020federated} or personalization strategies~\cite{li2021fedbn,sun2023fedperfix}, while segmentation-oriented federations adopt U-Net variants~\cite{wang2023feddp,jiang2023iop,jiang2023fair} and redesign local optimization to accommodate pixel-wise supervision. This one-fit-one design philosophy hinders reusability: switching tasks frequently necessitates re-engineering the FL pipeline instead of reusing a universal federated recipe. Furthermore, task adaptation is tightly entangled with model architecture decisions, substantially increasing engineering friction and limiting the practical scalability of FL deployment. However, even if such task-level rigidity is addressed, real-world practices must also confront a more fundamental challenge rooted in heterogeneous imaging modalities across institutions.

\begin{figure}[tbh]
    \centering
    \includegraphics[width=0.485\textwidth]{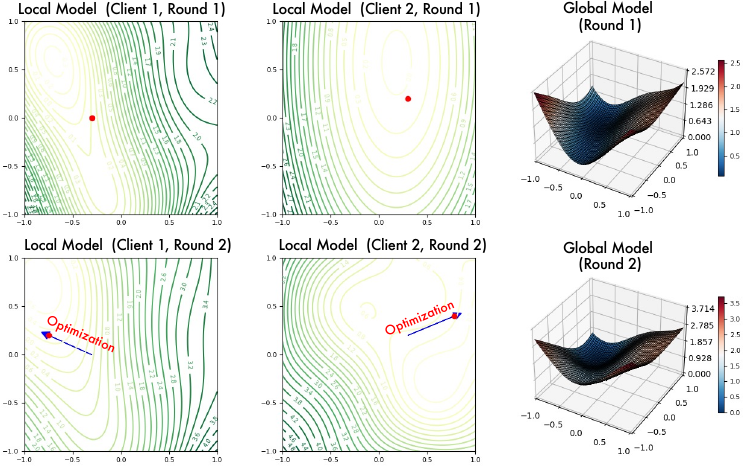}
    \caption{Optimization dynamics under heterogeneous imaging modalities, where MRI in Client 1 and Microscope in Client 2.}
    \label{fig:intro_fig_gra}
    \vspace{-6pt}
\end{figure}

While prior FL work predominantly investigates statistical heterogeneity arising from sampling bias~\cite{lu2023fedclip,chen2023prompt,chen2024federated,zhang2023fedcp,zhang2023fedala}, a more severe and largely under-explored challenge arises from heterogeneous imaging modalities across institutions, even when the clinical task is fixed within a federation. Hospitals may provide MRI, CT, PET, SPECT, or pathology images due to equipment availability or diagnostic preference. Such modality discrepancies exacerbate non-IID distributions and introduce fundamentally different feature statistics, visual textures, and frequency characteristics. Consequently, local models optimize against diverging loss landscapes; during aggregation, the global model is pulled toward conflicting minima. \textbf{Fig.~\ref{fig:intro_fig_gra}} visualizes this phenomenon: different modalities yield distinct local loss basins, causing clients to descend toward disparate minima. After aggregation, the global model shifts toward a suboptimal saddle-like region, leading to slower convergence and oscillatory behavior across communication rounds.

These observations raise a key question: \emph{Can we build a FL framework that \textbf{(i)} reuses a single optimization pipeline across different medical tasks without re-engineering, and \textbf{(ii)} remains robust to modality-induced discrepancies even when the clinical task is fixed within a federation?} Achieving both would drastically simplify real-world deployment by decoupling federation design from model selection and task ontology, while yielding consistent optimization behavior under heterogeneous imaging modalities.

To address these challenges, we propose \textbf{\texttt{OmniFM}}, an omni-purpose, modality- and task-agnostic FL framework for medical image analysis. As shown in \textbf{Fig.~\ref{fig:intro_fig}}, \textbf{\texttt{OmniFM}} maintains a unified, reusable FL optimization pipeline, allowing different tasks to reuse the same training mechanism without re-engineering. Its core design mitigates modality heterogeneity through heterogeneous frequency-aware reasoning, enabling stable and broadly applicable federated training across diverse medical imaging modalities. Specifically, \textbf{\texttt{OmniFM}} integrates three complementary components: (1) Global Spectral Knowledge Retrieval, which retrieves global spectral prototypes from a server-side knowledge bank to enrich local spectral embeddings; (2) Embedding-wise Cross-Attention Fusion, which injects retrieved global spectral context into backbone representations via cross-attention; and (3) Prefix–Suffix Spectral Prompting, which augments token sequences with learnable prefix and suffix prompts to enhance representation consistency. Additionally, a Spectral-Proximal Alignment term regularizes local updates, mitigating optimization drift caused by modality discrepancies. \textbf{\texttt{OmniFM}} unifies classification, segmentation, super-resolution, VQA, and multimodal fusion within a single framework, maintaining robustness and reusability across heterogeneous imaging modalities. The key contributions are summarized below:
\begin{itemize}
    \item We formulate the challenge of building a general-purpose FL framework that supports multiple medical image tasks and remains robust to heterogeneous modalities, a setting rarely addressed in prior work of FL in medical image.
    \item We propose \textbf{\texttt{OmniFM}}, an omni-purpose, modality- and task-agnostic FL framework that reuses a single optimization pipeline across runs. OmniFM can instantiate different model to support classification, segmentation, super-resolution, visual question answering, and multimodal fusion without altering the underlying FL mechanism.
    \item Guided by frequency-domain insights, we propose Global Spectral Knowledge Retrieval, Embedding-wise Cross-Attention, and Prefix–Suffix Spectral Prompting within OmniFM to inject informative context into local representations, while Spectral-Proximal Alignment term mitigates modality-induced optimization drift.
    \item Experiments on real-world medical datasets demonstrate that \textbf{\texttt{OmniFM}} outperforms advanced baseline across diverse tasks and modality configurations, covering both training from scratch and fine-tuning settings.
\end{itemize}

\section{Related Work}

\paragraph{Heterogeneous Federated Learning.} FL enables decentralized model training across distributed clients without sharing raw data~\citep{mcmahan2017communication,chen2025federated}. A key challenge is statistical heterogeneity (non-IID), which degrades performance due to distributional shifts across clients. Early approaches reduce client–server divergence through alignment or regularization, but generally assume mild heterogeneity and shared feature spaces that break down in diverse real-world domains. To address this, personalized FL (PFL) tailors models to individual clients using strategies such as regularization-based decomposition~\citep{hanzely2020lower,li2021ditto}, partial model sharing~\cite{chen2023prompt,li2021fedbn,collins2021exploiting,chenfedal}, adaptive aggregation~\citep{zhang2020personalized,chenfedal}, or meta-learning~\citep{fallah2020personalized}. However, most PFL methods remain task-specific (mostly optimization for classification) and tightly coupled to client architectures, limiting their ability to support multiple tasks under a unified federated pipeline.

\begin{figure*}[tbh]
    \centering
    \includegraphics[width=.95\textwidth]{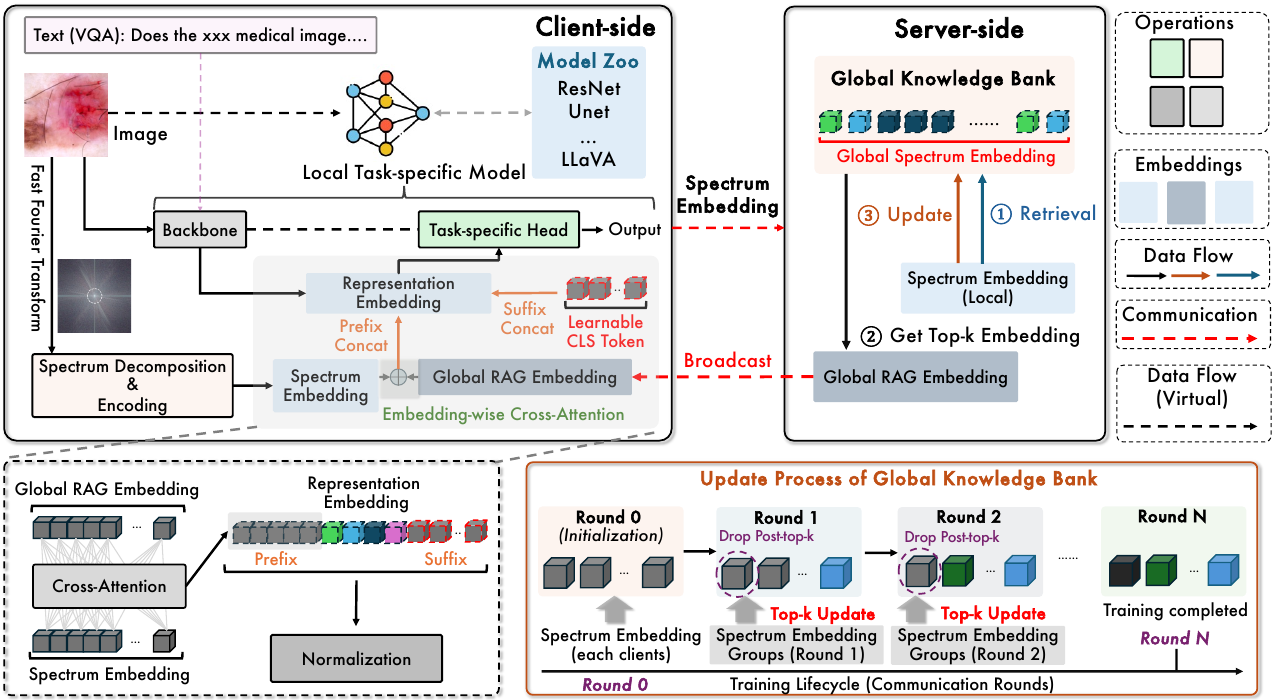}
    \caption{Overview of our \textbf{\texttt{OmniFM}}. Client-side image feature embedding and FFT-derived spectral embeddings are fused with retrieved global spectral prototypes via embedding-wise cross-attention and prefix-suffix prompting. Spectral embeddings are periodically uploaded to update a global knowledge bank, while a spectral-proximal alignment term mitigates modality-induced drift during optimization.}
    \label{fig:framework_overall}
    \vspace{-12pt}
\end{figure*}

\vspace{-8pt}
\paragraph{Federated Medical Imaging Learning.} FL has become a promising paradigm for privacy-preserving, multi-institutional medical imaging analysis~\cite{feng2025taming}. Federated medical imaging learning (FMIL) extends FL to tasks such as classification~\cite{ren2024federated,chen2023interpretable,feng2026visual,wang2024ensemble}, segmentation~\cite{miao2023fedseg,jiang2023fair,wang2023feddp}, and image fusion. However, differences in patient populations, scanner hardware, and acquisition protocols introduce severe non-IID challenges. Moreover, FL frameworks developed for natural images often fail to generalize to medical data due to fundamental differences in feature distributions and representational structures. Existing FMIL methods are typically task-specific~\cite{chen2025restyled,jiang2023fair,liu2024fedfms}, requiring customized optimization strategies for each backbone, which hinders reusability and scalability. Most also assume homogeneous imaging modalities, neglecting the more complex and realistic cross-modality setting that exacerbates representation divergence and aggregation drift. Our \textbf{\texttt{OmniFM}} addresses these limitations by introducing a unified, reusable FL protocol that remains robust across diverse tasks and heterogeneous modalities.

\section{Methodology}
\paragraph{Problem Formulation.} We consider a personalized FL setup involving $K$ distributed medical institutions (clients), indexed by $k \in \{1,\dots,K\}$. Each client maintains a private dataset $\mathcal{D}_k = \{(x_{k,i}, y_{k,i})\}_{i=1}^{n_k}$, where $x_{k,i}$ denotes a medical image sampled from one or more imaging modalities $\mathcal{M}_k \subseteq \mathcal{M}$ (e.g., MRI, CT, PET, pathology), and $y_{k,i}$ represents task-specific supervision. Modalities across clients may be partially overlapping or completely disjoint:
\begin{equation}
    \mathcal{M}_k \neq \mathcal{M}_j~\text{or}~\mathcal{M}_k \cap \mathcal{M}_j \neq \emptyset,\quad \forall k \neq j,
\end{equation}
inducing heterogeneous visual features and frequency spectra across local datasets. We define a personalized model $\theta_k$ for each client, and the global goal is to collaboratively learn a population of models $\{\theta_k\}$ while preventing raw data sharing. The optimization problem can be formulated as:
\begin{equation}
    \min_{\{\theta_k\}}~\sum_{k=1}^{K} \frac{n_k}{N} F_k(\theta_k), \quad N = \sum_{k=1}^{K} n_k,
\end{equation}
where $F_k(\theta_k) = \mathbb{E}_{(x,y)\sim\mathcal{D}_k}[\mathcal{L}(f_{\theta_k}(x), y)]$ denotes the expected loss on client $k$. To promote collaborative learning, these personalized models are implicitly coupled through server-side aggregation and shared components. We support two categories of downstream tasks. The single-modality setting involves pure medical images for classification, segmentation, super-resolution, and image fusion. The multi-modality setting incorporates auxiliary textual inputs $t$, where paired samples $(x_{k,i}, t_{k,i})$ are used for visual question answering or multimodal reasoning. Therefore, our goal is to develop a single federated optimization pipeline that supports diverse medical imaging tasks without re-engineering and remains robust to modality-induced heterogeneity. \textbf{\texttt{OmniFM}} achieves this through frequency-aware reasoning mechanisms, enabling consistent performance across heterogeneous imaging modalities.

\subsection{Framework Overview}
\begin{figure}[tbh]
    \centering
    \includegraphics[width=0.485\textwidth]{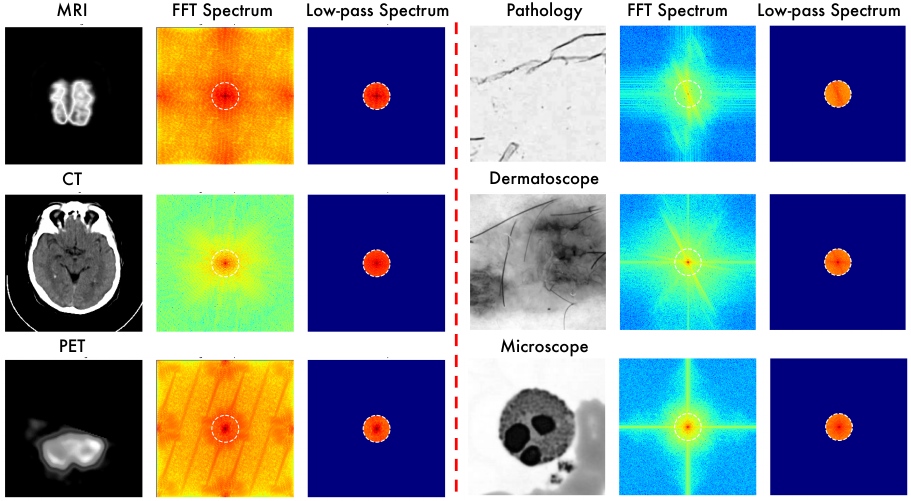}
    \caption{Spectrum from different imaging modalities.}
    \label{fig:fft_insights}
    \vspace{-20pt}
\end{figure}

\begin{tcolorbox}[
    colback=violet!5,
    colframe=violet!20,
    arc=1mm,
    boxrule=0pt,
    leftrule=0pt,
    rightrule=0pt,
    toprule=0pt,
    bottomrule=0pt,
    boxsep=1pt,
    left=1.5pt,
    right=1.5pt,
    top=1.5pt,
    bottom=1.5pt,
    enhanced,
]
\textbf{Insights from Spectrum Observation.}
\emph{Medical imaging modalities differ widely in factors like acquisition physics, spatial statistics; however, their low-frequency spectral structures remain strikingly consistent across modalities (\textbf{Fig.\ref{fig:fft_insights}}). Let $\mathcal{F}(x)$ denote the magnitude spectrum of an image x, and let $\mathcal{P}_{\mathrm{LP}}(\cdot)$ extract its low-frequency band. For different modalities, we empirically observe}
\[
\left\| \mathcal{P}_{\mathrm{LP}}(\mathcal{F}(x_i)) - \mathcal{P}_{\mathrm{LP}}(\mathcal{F}(x_j)) \right\|_2
\;\ll\; 
\left\| \mathcal{F}(x_i) - \mathcal{F}(x_j) \right\|_2,
\]\emph{where $\forall x_i \in \mathcal{M}_a$,$x_j \in \mathcal{M}_b,\, a \ne b$ indicating that modality-specific variability is predominantly concentrated in higher frequencies, while low-pass components encode coarse anatomical structures that are shared across modalities. This suggests that $\mathcal{P}_{\mathrm{LP}}(\mathcal{F}(x))$ can serve as a modality-invariant structural prior for representation learning. Motivated by this insight, \textbf{\texttt{OmniFM}} incorporates frequency-aware mechanisms that inject global spectral context and regularize spectral similarity during optimization, thereby mitigating modality-induced representation divergence and stabilizing federated aggregation under heterogeneous imaging modalities.}
\end{tcolorbox}

\noindent \textbf{Fig.~\ref{fig:framework_overall}} presents the architecture of \textbf{\texttt{OmniFM}}. Each client extracts representation embeddings through a backbone and obtains local spectral embeddings via FFT-based spectrum encoding. The local spectral embedding queries a server-side global knowledge bank to retrieve top-$k$ global spectral prototypes, which are fused into backbone representations via embedding-wise cross-attention and prefix–suffix prompting before being processed by a task-specific head. Clients periodically upload spectral embeddings to refine the global knowledge bank across communication rounds. A spectral-proximal alignment term further suppresses modality-induced drift, ensuring stable aggregation under heterogeneous imaging modalities.

\subsection{Local Updating on Clients}
During each communication round $r$, following~\cite{zhang2023fedala}, we factorize the local model into a backbone $g_{\phi}$ and a task head $h_{\psi}$ to decouple representation learning from task-level specialization. Both components are retained locally, enabling clients to adapt to their own modality distributions without exposing full model parameters. Given an input sample $x \in \mathbb{R}^{H \times W \times C}$, the backbone produces a representation embedding: $\mathbf{r} = g_{\phi}(x) \in \mathbb{R}^{L \times d}$, where $L$ and $d$ denote token length and embedding dimension, respectively. For notational simplicity, we omit client-specific subscripts on parameters and embeddings; all quantities should be understood as local to each participating client. We focus primarily on image samples for clarity. In VQA involving text, we follow~\cite{liu2023visual}, mapping the text stream to the same downstream interface. Therefore, we omit explicit notation for text samples in subsequent equations. The representation $\mathbf{r}$ encodes semantic structure for the downstream task but remains sensitive to modality-induced statistical shifts, motivating the frequency-aware modules introduced below.

\vspace{-8pt}
\paragraph{Frequency-Domain Spectrum Encoding.} To extract modality-invariant structural cues, we first transform an input image into the frequency domain as $\mathbf{F} = \operatorname{FFT}(x)$ via fast fourier transform, and compute its magnitude spectrum. We then perform spectrum decomposition by applying a low-pass projection operator $\mathcal{P}_{\mathrm{LP}}$ as $\mathbf{S}_{\mathrm{LP}} = \mathcal{P}_{\mathrm{LP}}(|\mathbf{F}|),$ which preserves coarse anatomical structures while attenuating modality-specific high-frequency variations. Next, we encode $\mathbf{S}_{\mathrm{LP}}$ through a Spectral Tokenization Module, which consists of a lightweight spectral mixing operator followed by learnable projections and global spectral pooling. The process can be formulated as:
\begin{equation}
    \mathbf{s} =
\operatorname{Pool}\Big(
\sigma( \mathbf{W}_2 \cdot \sigma(\mathbf{W}_1 \cdot \operatorname{FreqMix}(\mathbf{S}_{\mathrm{LP}})))
\Big),
\end{equation}
where $\operatorname{FreqMix}$ aggregates frequency responses across spectral channels. The resulting spectral token $\mathbf{s}$ captures modality-invariant anatomical priors, serving as a stable anchor for subsequent retrieval and fusion. To prevent spectral collapse, we normalize $\mathbf{s}$ to unit length $\mathbf{s} \leftarrow \frac{\mathbf{s}}{\|\mathbf{s}\|_2}.$

\vspace{-8pt}
\paragraph{Retrieval-Augmented Embedding Fusion.} To complement locally observed modality characteristics, we retrieve global spectral priors from a server-side knowledge bank $\mathcal{K}^{(r)} = \{\mathbf{s}^{(1)}, \dots, \mathbf{s}^{(M)}\}$, which accumulates spectral embeddings across communication rounds. Given a local spectral token $\mathbf{s}$, we compute cosine similarities $\alpha_i = \cos(\mathbf{s}, \mathbf{s}^{(i)})$ and select the top-$k$ most similar prototypes, yielding the Global RAG Embedding $\mathbf{S}_{g} \in \mathbb{R}^{k \times d}$. These retrieved prototypes provide cross-modality spectral context beyond what can be learned from an individual client, particularly beneficial under imbalanced or rare modality distributions. We fuse this global spectral context into the backbone representation $\mathbf{r} \in \mathbb{R}^{L \times d}$ through embedding-wise cross-attention \textbf{(ECA)}. Specifically, we form queries $\mathbf{Q} = \mathbf{r}\mathbf{W}_Q$ and keys/values $\mathbf{K} = \mathbf{S}_g\mathbf{W}_K, \mathbf{V} = \mathbf{S}_g\mathbf{W}_V$, where $\mathbf{W}_Q, \mathbf{W}_K, \mathbf{W}_V \in \mathbb{R}^{d \times d_h}$ are learnable projections. The ECA-fusion output is:
\begin{equation}
    \mathbf{Z} = \operatorname{Softmax}\left(\frac{\mathbf{Q}\mathbf{K}^\top}{\sqrt{d_h}}\right)\mathbf{V}.
\end{equation}
This operation conditions local tokens on global low-frequency priors, biasing representations toward modality-invariant anatomy over modality-specific texture.
\vspace{-8pt}
\paragraph{Prefix–Suffix Spectral Prompting.} The fused spectral tokens $\mathbf{Z}$ encapsulate globally consistent, frequency-aware cues aligned to local representations. To further condition downstream inference, we introduce Prefix–Suffix Spectral Prompting \textbf{(PSP)}, which injects both global and personalized priors into the token sequence. Specifically, $\mathbf{Z}$ is prepended as prefix tokens to the backbone features $\mathbf{r}$, while learnable, client-specific CLS tokens $\mathbf{c}\in\mathbb{R}^{p\times d}$ are appended as suffix prompts: $\mathbf{r}’ = [\,\mathbf{Z} \,\|\, \mathbf{r} \,\|\, \mathbf{c}\,],$ where $|$ denotes token concatenation. The prefixes bias representations toward modality-invariant, low-frequency structure, whereas the suffixes capture high-level adaptations tailored to each institution’s distribution. This dual prompting scheme jointly enhances inter-client consistency while preserving local specialization. The resulting token sequence $\mathbf{r}’$ is then fed into the task-specific head $h_{\psi}$ to generate predictions robust to heterogeneous imaging modalities.

\vspace{-10pt}
\paragraph{Spectral-Proximal Alignment in Optimization.} Although prompting provides modality-invariant cues, local updates may still drift toward modality-specific minima. To address this, we introduce Spectral-Proximal Alignment \textbf{(SPAlign)}, which enforces proximity between the local spectral embedding $\mathbf{s}$ and the centroid of the retrieved global prototypes. Let $\bar{\mathbf{s}}_g = \frac{1}{k}\sum_{\mathbf{s}’ \in \mathbf{S}_g} \mathbf{s}’$ denote the prototype barycenter. We formulate the alignment objective as: 
\begin{equation}
    \mathcal{L}_{\text{align}}(\mathbf{s}, \mathbf{S}_g)
=  \left\| \mathbf{s} - \bar{\mathbf{s}}_g \right\|_2^2,
\end{equation}
which regularizes the spectral geometry of local representations while preserving flexibility in the spatial domain. The complete local objective becomes as below:
\begin{equation}
    \min_{\phi, \psi}~\mathcal{L}_{\text{task}}\!\left(h_{\psi}\big([\mathbf{Z}||\mathbf{r}||\mathbf{c}]\big), y\right)
\;+\;\lambda\mathcal{L}_{\text{align}},
\end{equation}
where $\lambda > 0$ controls the strength of spectral regularization, and $\mathcal{L}_{\text{task}}$ denotes the standard objective for each task. By anchoring frequency-domain embeddings to globally consistent low-frequency priors, SPAlign mitigates modality-induced optimization drift and stabilizes aggregation. 
\begin{table*}[tbh]
\centering
\caption{Results on federated cross-modality classification under three heterogeneous scenarios (more details in Appendix). We employ ResNet-18~\cite{he2016deep} as the local backbone, train for 100 rounds with 5 local epochs per round. \boldres{Bold}: the best, \secondres{underline}: the second best.}
\vspace{-8pt}
\setlength{\tabcolsep}{3.5pt}
\resizebox{.925\textwidth}{!}{
\begin{tabular}{lcccccc|cccccc|cccccc}
\toprule
\multirow{4}{*}{\textbf{Method}}
& \multicolumn{6}{c}{\textbf{Scenario 1 (Hard)}}
& \multicolumn{6}{c}{\textbf{Scenario 2 (Medium)}}
& \multicolumn{6}{c}{\textbf{Scenario 3 (Extra)}} \\
\cmidrule(lr){2-7} \cmidrule(lr){8-13} \cmidrule(lr){14-19}

& \multicolumn{3}{c}{\textbf{Acc (\%)}}
& \multicolumn{3}{c}{\textbf{F1-Score}}
& \multicolumn{3}{c}{\textbf{Acc (\%)}}
& \multicolumn{3}{c}{\textbf{F1-Score}}
& \multicolumn{3}{c}{\textbf{Acc (\%)}}
& \multicolumn{3}{c}{\textbf{F1-Score}} \\
\cmidrule(lr){2-4} \cmidrule(lr){5-7} \cmidrule(lr){8-10} \cmidrule(lr){11-13} \cmidrule(lr){14-16} \cmidrule(lr){17-19}

& {20\%} & {50\%} & {100\%}
& {20\%} & {50\%} & {100\%}
& {20\%} & {50\%} & {100\%}
& {20\%} & {50\%} & {100\%}
& {20\%} & {50\%} & {100\%}
& {20\%} & {50\%} & {100\%} \\
\midrule

FedAvg~\cite{mcmahan2017communication} & 56.31 & 67.31 & 84.21 & 0.234 & 0.380 & 0.474 & 56.44 & 67.11 & 84.49 & 0.229 & 0.334 & 0.460 & 24.24 & 56.84 & \secondres{99.49} & 0.165 & 0.522 & 0.977 \\
FedProx~\cite{li2020federated} & 27.04 & 58.79 & 93.92 & 0.060 & 0.293 & 0.491 & 30.17 & 38.33 & 30.14 & 0.119 & 0.124 & 0.131 & 24.07 & 55.79 & 97.53 & 0.130 & 0.369 & 0.587 \\
Ditto~\cite{li2021ditto} & 28.67 & 65.31 & \secondres{97.56} & 0.096 & 0.385 & 0.649 & 30.23& 69.23 & 96.11 & 0.123& 0.486 & 0.820 & 25.09 & 58.66 & 99.24 & 0.170 & 0.447 & 0.931\\
SCAFFOLD~\cite{karimireddy2020scaffold} & 24.09 & 16.56 & 19.85 & 0.046 & 0.063 & 0.082 & 24.04 & 18.01 & 20.59 & 0.053 & 0.067 & 0.078 & 52.31 & 49.99 & 99.41 & 0.241 & 0.180 & 0.846\\
MOON~\cite{li2021model} & 28.46 & 50.99 & 97.52 & 0.085 & 0.289 & 0.644 & 29.65 & 23.37 & 22.90 & 0.136 & 0.085 & 0.063 & 19.07 & 56.81 & 98.99 & 0.154 & 0.528 & 0.881 \\
GPFL~\cite{zhang2023gpfl} & 57.78 & 62.20 & 52.88 & 0.303 & 0.319 & 0.208 & 68.04 & 56.47 & 54.98 &0.284 & 0.313 & 0.254 & 68.09 & 54.52 & 54.72& 0.338 & 0.436& 0.505\\
FedDyn~\cite{acar2021federated} & 64.37 & 87.85 & 94.99 & 0.315 & 0.460 & 0.554 & 74.92 & \secondres{88.94} & \secondres{94.99} & 0.283 & 0.508 & 0.554 & 73.29 & 86.12 & 90.46 & 0.387 & 0.512 & 0.550 \\
FedRep~\cite{collins2021exploiting} & 83.52 & 89.01 & 93.46 & 0.395 & 0.432 & 0.487 & 82.43& 88.09 & 92.54 & 0.365 & 0.444& 0.525 & 85.10 & 96.02 & 98.06 & 0.624 & 0.836 & 0.908 \\
FedPer~\cite{arivazhagan2019federated} & \secondres{84.47} & \secondres{91.02} & \secondres{92.57} & \secondres{0.461} & \secondres{0.621} & \secondres{0.665} & 83.51 & 91.02 & 92.80 & 0.409 & \secondres{0.621} & 0.654 & \secondres{91.81} & \secondres{98.76} & 99.41 & \secondres{0.848} & \secondres{0.911} & \boldres{0.980} \\
\midrule
\textbf{\texttt{OmniFM} (Ours)}
& \boldres{96.85} & \boldres{97.71} & \boldres{97.82}
& \boldres{0.589} & \boldres{0.663} & \boldres{0.668}
& \boldres{96.27} & \boldres{96.88} & \boldres{97.01}
& \boldres{0.576} & \boldres{0.657} & \boldres{0.666}
& \boldres{98.85} & \boldres{99.30} & \boldres{99.77}
& \boldres{0.959} & \boldres{0.971} & \secondres{0.978} \\
\bottomrule
\end{tabular}}
\vspace{-4pt}
\label{tab:classification}
\end{table*}

\begin{table*}[tbh]
    \centering
    \caption{Results on super-resolution with three independent clients across three scales. \textbf{Bold}: the best, \underline{Underline}: the second best.}
    \vspace{-8pt}
    \setlength{\tabcolsep}{2pt}
   \setlength{\extrarowheight}{3pt}
    \resizebox{.93\textwidth}{!}{
    \begin{tabular}{lcccccc|ccccc|ccccc|ccccc}
    \toprule

    \multirow{3}{*}{\textbf{Scale}} & \multirow{3}{*}{\textbf{Client}} & \multicolumn{10}{c}{\textbf{Scenario 1}} & \multicolumn{10}{c}{\textbf{Scenario 2}} \\
    \cmidrule(lr){3-12} \cmidrule(lr){13-22}
    
    & & \multicolumn{5}{c}{\textbf{PSNR} $\mathbf{\uparrow}$} & \multicolumn{5}{c}{\textbf{SSIM} $\mathbf{\uparrow}$} &
    \multicolumn{5}{c}{\textbf{PSNR} $\mathbf{\uparrow}$} &
    \multicolumn{5}{c}{\textbf{SSIM} $\mathbf{\uparrow}$} \\
    \cmidrule(lr){3-7} \cmidrule(lr){8-12} \cmidrule(lr){13-17} \cmidrule(lr){18-22}
    
    & & \textbf{\texttt{OmniFM}} & FedAvg & FedProx & FedPer & Local &\textbf{\texttt{OmniFM}} & FedAvg & FedProx & FedPer & Local
    &\textbf{\texttt{OmniFM}} & FedAvg & FedProx & FedPer & Local
    &\textbf{\texttt{OmniFM}} & FedAvg & FedProx & FedPer & Local\\
    \midrule
        
        \multirow{5}{*}{$\mathbf{\times 2}$} 
        & Client 1 & \boldres{42.93} & 35.90 & 20.68 & \secondres{40.24} & 36.97&\boldres{0.9784} & 0.9526 & 0.5426 & \secondres{0.9537} &0.9115 & \boldres{43.11} & 41.67 & 24.82 &41.42 & 40.38& \boldres{0.9811} & 0.9399 & 0.8230 & 0.9610 &\secondres{0.9583}\\
        & Client 2 & \boldres{41.95} & 36.07 & 22.42 & \secondres{39.32} &35.09 &\boldres{0.9634} & 0.9444 & 0.5462 & \secondres{0.9478}& 0.9231 & \boldres{41.95} & 41.70 & 24.09 & 41.98& 42.23& \boldres{0.9690} & 0.9582 & 0.8152 &0.9631 &  \secondres{0.9651}\\
        & Client 3 & \boldres{41.75} & 35.87 & 22.44 & \secondres{39.00} & 34.81&\boldres{0.9661} & 0.9543 & 0.5513 & \secondres{0.9567} &0.9533 & \boldres{41.85} & 41.12 & 24.20 & 39.22& 34.92& 0.9634 & \boldres{0.9661} & 0.8076 & \secondres{0.9527} & 0.9472\\
        \cmidrule(lr){2-22}
        & \bf Avg.    & \boldres{42.21} & 35.95 & 21.84 & \secondres{39.52} & 35.63& \boldres{0.9693} & 0.9505 & 0.5467 & \secondres{0.9527}& 0.9293 & \boldres{42.30}& 41.50 & 24.37 & 40.87& 39.17& 0.9711 & 0.9548 & 0.8152 & 0.9590 &0.9569 \\
        \midrule
        
        \multirow{5}{*}{$\mathbf{\times 4}$}
        & Client 1 & \boldres{36.33} & 30.18 & 9.10  & \secondres{30.52} &30.97 &\boldres{0.8990} & 0.8659 & 0.0971 & 0.8724& \secondres{0.8891} & \boldres{37.73} & \secondres{34.50} &  9.87 & 33.81&  32.77& \boldres{0.9335} & \secondres{0.9164} & 0.1180 &  0.9126& 0.8966\\
        & Client 2 & \boldres{34.81} & \secondres{30.64} & 9.40  & 30.30 &29.83 &\secondres{0.8569} & 0.8518 & 0.1205& 0.8542& \boldres{0.8614} & \boldres{38.52} & 34.28 & 9.88 &34.73 & \secondres{35.45} & \boldres{0.9507} & 0.9165 & 0.1088 & 0.9175& \secondres{0.9246}\\
        
        & Client 3 & \boldres{34.30} & \secondres{30.32} & 9.26  & 30.03 & 29.45
        &0.8483 & \boldres{0.8597} & 0.1126 & \secondres{0.8555} & 0.8531 & \boldres{34.11} & \secondres{33.62} & 9.71 & 31.21&  29.38& \secondres{0.8576}& \boldres{0.9048} & 0.1078 & 0.8057& 0.8567\\
        \cmidrule(lr){2-22}
        & \bf Avg.     & \boldres{35.15} & \secondres{30.38} & 9.25  & 30.29 & 30.08&\boldres{0.8681} & 0.8592 & 0.1101 & \secondres{0.8607} & 0.8679 & \boldres{36.79} & \secondres{34.14} & 9.82 &33.25 & 32.54 &\boldres{0.9139} & \secondres{0.9125} & 0.1115 & 0.8786&  0.8927\\
        \midrule
        
        \multirow{5}{*}{$\mathbf{\times 8}$}
        & Client 1 & \boldres{30.12} & \secondres{28.96} & 11.26 & 29.11 & 29.28 &\boldres{0.7742} & 0.7534 & 0.1350 & 0.7532 & \secondres{0.7605} & \boldres{31.24} & 30.70 & 12.18 & \secondres{30.52} & 30.43 & \boldres{0.7958} & \secondres{0.7847} & 0.1609 &0.7807 &0.7791\\
        
        & Client 2 & \boldres{30.27} & \secondres{29.81} & 12.48 & 29.45 &29.42 &\boldres{0.7644} & 0.7596 & 0.1687 & \secondres{0.7600} & 0.7482 & \boldres{31.26}& 30.12 & 11.95 & 30.24 & \secondres{30.29} & \boldres{0.8084} & 0.7792 &  0.1468&0.7808 & \secondres{0.7818}\\
        
        & Client 3 & \boldres{29.68} & \secondres{29.32} & 11.89 & 28.91 &28.75
        &0.7520 & \boldres{0.7552} & 0.1558 & \secondres{0.7521} & 0.7412 & \secondres{29.41}&  \boldres{29.70} & 11.72 &  28.81& 28.57 &\secondres{0.7508} & \boldres{0.7659} & 0.1445 & 0.7457& 0.7389\\
        \cmidrule(lr){2-22}
        & \bf Avg.     & \boldres{30.02} & \secondres{29.36} & 11.88 & 29.15 & 29.14 &\boldres{0.7636} & \secondres{0.7557} & 0.1531 & 0.7551 & 0.7500 &\boldres{30.64} & \secondres{30.17} &  11.95 &29.86 & 29.76 & \boldres{0.7850}& \secondres{0.7766} & 0.1507 & 0.7691& 0.7666\\
        \midrule
        \multicolumn{2}{c}{\textbf{Average}} & \boldres{35.79} & 31.90 & 14.32  & \secondres{32.99}  &31.62 & \boldres{0.8670} & 0.8551 & 0.2700 & \secondres{0.8562} &  0.8490& \boldres{36.58} & \secondres{35.27} & 15.38& 34.66 & 33.82 & \boldres{0.8900}& \secondres{0.8813} & 0.3591 & 0.8689 & 0.8720\\
        \bottomrule
    \end{tabular}}
    \label{tab:superresolution}
    \vspace{-10pt}
\end{table*}
\subsection{Global Spectral Knowledge Retrieval}
To achieve Global Spectral Knowledge Retrieval (GSKR), the server maintains a global knowledge bank $\mathcal{K}^{(r)} \subset \mathbb{R}^d$, which stores spectral embeddings accumulated across communication rounds. Each participating client uploads only its local spectrum embedding $\mathbf{s}_k^{(r)}$, without sharing backbone or head parameters, reducing communication overhead and preserving privacy. Upon request, the server performs a retrieval–augmentation routine consisting of three steps (Fig.~\ref{fig:framework_overall}), including (1) compute cosine similarities as $\alpha_i = \cos(\mathbf{s}_k^{(r)}, \mathbf{s}^{(i)})$, (2) select the top-$k$ prototypes as $\mathbf{S}g = \operatorname{Top}\text{-}k\big(\mathcal{K}^{(r)}, \{\alpha_i\}\big)$, and (3) insert $\mathbf{s}_k^{(r)}$ back into the bank under a bounded spectral constraint:
\begin{equation}
    \mathcal{K}^{(r+1)} = \Pi_{\mathbb{B}(\rho)}\!\Big(\mathcal{K}^{(r)} \cup \{\mathbf{s}_k^{(r)}\}_{k=1}^{K_r}\Big),
\end{equation}
where $\Pi_{\mathbb{B}(\rho)}(\cdot)$ denotes projection onto a Hilbert ball $\mathbb{B}(\rho)=\{\mathbf{s}\mid \|\mathbf{s}\|_2\le \rho\}$, preventing unbounded drift in the spectral embedding geometry. Finally, aggregated backbone parameters and updated bank statistics are broadcast to clients for the next round of frequency-aware optimization.
\vspace{-10pt}
\paragraph{Knowledge Bank Update.} The knowledge bank evolves progressively over communication rounds. At initialization (Round 0), spectral embeddings uploaded from all clients form the seed set $\mathcal{K}^{(0)} = \{\mathbf{s}_k^{(0)}\}_{k=1}^{K}.$ At round $r>0$, newly received embeddings update the bank via:
\begin{equation}
    \mathcal{K}^{(r+1)} = \operatorname{Prune}\!\Big(\mathcal{K}^{(r)} \cup \{\mathbf{s}_k^{(r)}\}_{k=1}^{K_r}\Big),
\end{equation}
where the pruning operator removes prototypes that consistently fall outside top-$k$ retrieval frequency $\operatorname{Prune}(\mathbf{s}^{(i)}) = \emptyset \quad \text{if} \quad \operatorname{freq}(\mathbf{s}^{(i)}) < \delta$. This post–top-$k$ filtering preserves prototype diversity, enabling $\mathcal{K}^{(r)}$ to converge to a compact, modality-balanced set of spectral anchors that stabilize aggregation under cross-modality heterogeneity.

\section{Main Results}
In this section, we comprehensively evaluate \textbf{\texttt{OmniFM}} on heterogeneous federated medical imaging tasks, spanning classification, segmentation, super-resolution, multimodal fusion, and medical visual question answer. Experiments show that \textbf{\texttt{OmniFM}} consistently achieves superior performance across fine-tuning and training-from-scratch setups.

\subsection{Cross-modality Classification}
\vspace{-2.7pt}
\textbf{Datasets \& Setups.} We construct a cross-modality classification dataset using multiple subsets of MedMNIST-v2~\cite{yang2023medmnist}, covering Colon Pathology, Dermatoscope, and Blood Cell Microscopy. The dataset contains 24 classes and approximately 110k samples, exhibiting substantial modality-induced heterogeneity. Following~\cite{zhang2023fedala}, we simulate non-iid via Dirichlet partitioning with coefficient $\gamma \in \{0.1, 0.5\}$ (medium and hard), and further introduce an extra-hard modality-heterogeneous setup. Performance is evaluated under participation ratios of $\{20\%, 50\%, 100\%\}$.

\noindent \textbf{Results.} The results are shown in Table~\ref{tab:classification}. Across all three heterogeneous scenarios and varying client participation ratios (20\%, 50\%, and 100\%), our \textbf{\texttt{OmniFM}} consistently achieves the highest Accuracy and F1-Score, outperforming strong personalized and heterogeneity-aware FL baselines. Notably, its performance remains stable even under the hardest setting, demonstrating strong robustness to severe cross-modality drift and partial client participation.

\subsection{Super-resolution}
\textbf{Dataset \& Setups.} We used BreaKHis dataset~\cite{spanhol2015dataset} for federated superresolution tasks, it consist of breast cancer histopathological image with eight classes. To simulate real-world heterogeneity for BreaKHis, we construct two non-IID dataset from the original dataset covering three super-resolution scale $\{\times2, \times4, \times8\}$ via performance corresponding downscaling~\cite{xie2023shisrcnet}. Dataset setup in Appendix.

\noindent \textbf{Results.} The intra-modality super-resolution results in Table~\ref{tab:superresolution} show that \textbf{\texttt{OmniFM}} consistently achieves the highest PSNR and SSIM across all heterogeneity scenarios and upscaling factors. The performance advantage over both personalized (FedPer) and heterogeneity-mitigating (FedProx/FedDyn) baselines widens as modality heterogeneity and scaling difficulty increase, underscoring the robustness of our frequency-aware design under severe distribution shifts.

\subsection{Visual Question Answer}
\textbf{Dataset \& Setups.} We construct three federated medical VQA settings to comprehensively evaluate the effectiveness of \textbf{\texttt{OmniFM}} using SLAKE~\cite{liu2021slake}, VQA-RAD~\cite{lau2018dataset}, VQA-Med 2019~\cite{ben2019vqa}, VQA-Med 2020~\cite{ben2021overview}, and VQA-Med 2021~\cite{ben2021overview}. In \textbf{Task 1}, we benchmark performance under varying heterogeneity and LLaVA fine-tuning strategies by using only SLAKE, removing question types with fewer than 20 samples, and performing modality-based partitioning to generate IID and non-IID splits across three clients. In \textbf{Task 2}, we assess robustness under complete modality heterogeneity by constructing eight modality-specific clients. In \textbf{Task 3}, we evaluate mixed-modality heterogeneity under different parameter-efficient tuning techniques by treating each dataset as an independent client, resulting in five heterogeneous clients. Details are in Appendix.

\begin{table}[tbh]
    \centering
    \caption{Results on Task 1 under IID and non-IID. F-C and F-L denote connector-only and LLM-only tuning, respectively; LC and 2-stage represent joint one-stage and sequential connector–LLM tuning. Best are in \boldres{bold} and second-best are \secondres{underlined}. Following, we adopt LLaVA-1.5 with a CLIP ViT-B/32 visual encoder and LLaMA-3.2-3B text backbone, and train LLM components via LoRA~\cite{hu2021lora} ($r$ = 8, $\alpha$ = 32) for parameter-efficient fine-tuning.}
    \vspace{-8pt}
    \setlength{\tabcolsep}{2pt}
   \setlength{\extrarowheight}{2pt}
    \resizebox{.485\textwidth}{!}{
    \begin{tabular}{l*{4}{cc}}
        \toprule
        \multirow{2}{*}{\makecell[l]{\textbf{Method}}} & 
        \multicolumn{2}{c}{\textbf{F-C}} & 
        \multicolumn{2}{c}{\textbf{F-L}} & 
        \multicolumn{2}{c}{\textbf{F-CL}} & 
        \multicolumn{2}{c}{\textbf{F-2stage}} \\
        \cmidrule(lr){2-3} \cmidrule(lr){4-5} \cmidrule(lr){6-7} \cmidrule(lr){8-9} 
        & IID & Non-IID & IID & Non-IID & IID & Non-IID & IID & Non-IID \\
        \midrule
        FedAvg~\cite{mcmahan2017communication} & 0.783 & 0.775 & 0.806 & 0.802 & 0.823 & 0.827 & 0.811 & 0.814 \\
        FedProx~\cite{li2020federated} & 0.734 & 0.750 & 0.800 & 0.780 & 0.816 & 0.796 & 0.773 & 0.785 \\
        FedAdam~\cite{reddi2020adaptive} & 0.741 & 0.735 & 0.783 & 0.771 & 0.777 & 0.774 & 0.782 & 0.777 \\
        FedAvgM~\cite{hsu2019measuring} & 0.754 & 0.747 & 0.789 & 0.786 & 0.784 & 0.768 & 0.793 & 0.794 \\
        FedYogi~\cite{reddi2020adaptive} & 0.745 & 0.736 & 0.782 & 0.769 & 0.783 & 0.774 & 0.785 & 0.782 \\
        FedDyn~\cite{acar2021federated} & 0.795 &  0.783 & \secondres{0.812} & 0.800 & 0.820 & 0.827 & 0.809 & 0.816 \\
        FedPer~\cite{arivazhagan2019federated}  & \secondres{0.799} & \secondres{0.790} & 0.809 & \secondres{0.814} & \secondres{0.827} & \boldres{0.833} & \secondres{0.818} & \secondres{0.824} \\
        \midrule
        \textbf{\texttt{OmniFM}} & \boldres{0.812} & \boldres{0.800} & \boldres{0.810} & \boldres{0.827} & \boldres{0.834} & \secondres{0.831} & \boldres{0.832} & \boldres{0.840} \\
        \bottomrule
    \end{tabular}}
    \label{tab:task1_vqa}
    \vspace{-12pt}
\end{table}
\begin{table}[tbh]
    \centering
    \caption{Results on Task 2. C1: Computed Tomography, C2: Ultrasound, C3: Optical Coherence Tomography, C4: Fundus Photography, C5: Light Microscopy, C6: Histopathology, C7: Dermatoscopy, C8: Chest X-ray. Training protocol is consistent with Table~\ref{tab:task1_vqa}. \boldres{Bold}: the best, and \secondres{underline}: the second best.} 
    \vspace{-8pt}
    \setlength{\tabcolsep}{2pt}
   \setlength{\extrarowheight}{3pt}
    \resizebox{.475\textwidth}{!}{
    \begin{tabular}{lccccccccc}
        \toprule
        \bf Method & \multicolumn{8}{c}{\bf Task-Specific Performance} & \bf Avg. \\ 
        \cmidrule(lr){2-9}
        & C1 & C2 & C3 & C4 & C5 & C6 & C7 & C8 & \\
        \midrule
        FedAvg & 88.67 & 72.59 & 70.86 & 85.93 & 71.09 & 89.45 & 66.62 & 76.48 & 77.71 \\
        FedProx & 88.92 & 73.05 & 71.03 & 86.10 & 71.40 & 89.20 & 66.81 & 76.32 & 77.85 \\
        FedAdam & 89.10 & \secondres{74.22} & 71.58 & 85.70 & 71.85 & 89.80 & 67.20 & 76.90 & 78.17 \\
        FedDyn & 88.75 & 73.90 & 71.92 & \secondres{86.35} & 71.55 & \secondres{90.02} & 67.45 & 76.70 & 78.08 \\
        FedPer & \secondres{89.43} & 74.10 & \secondres{72.21} & 85.95 & \secondres{72.30} & 89.90 & \secondres{67.82} & \secondres{77.52} & \secondres{78.40} \\
        \midrule
        \textbf{\texttt{OmniFM}} & \boldres{90.02} & \boldres{75.55} & \boldres{72.83} & \boldres{87.21} & \boldres{73.10} & \boldres{90.25} & \boldres{68.15} & \boldres{78.05} & \boldres{79.27} \\
        \bottomrule
    \end{tabular}}
    \label{tab:vqa_task2}  
    \vspace{-10pt}
\end{table}

\noindent \textbf{Results.} As shown in Table~\ref{tab:task1_vqa}, \textbf{\texttt{OmniFM}} consistently outperforms all FL baselines under both IID and non-IID settings across all tuning strategies, with the largest gains observed in connector-LLM joint optimization. For Task 2 (Table~\ref{tab:vqa_task2}), where each client holds a distinct imaging modality, \textbf{\texttt{OmniFM}} achieves the highest average performance, demonstrating strong resilience to severe cross-modality discrepancies. On Task 3 (Table~\ref{tab:vqa_task3}), where each dataset acts as an independent mixed-modality client, \textbf{\texttt{OmniFM}} maintains performance advantages across all adapter configurations and PEFT regimes, highlighting robustness to dataset-level heterogeneity and improved generalization under diverse modality mixtures. These indicate that our frequency-aware strategies and alignment mechanisms effectively  enhance cross-modality compatibility in federated VQA.

\begin{table}[tbh]
    \centering
    \vspace{-7pt}
    \caption{Results on Task 3. Training protocols follow Table~\ref{tab:task1_vqa}. \boldres{Bold} indicates the best and \secondres{underline} denotes the second best.}
    \vspace{-8pt}
    \resizebox{.48\textwidth}{!}{
    \begin{tabular}{lcccccc}
        \toprule
        \textbf{Method} & \textbf{SLAKE} & \textbf{VM2019} & \textbf{VM2020} & \textbf{VM2021} & \textbf{VR} & \textbf{Average} \\
        \midrule
        \bf Full Tuning       \\
        \qquad $\triangleright$ \color{midgray}{FedAvg} & 77.45 & 67.25 & 15.06 & 22.00 & 41.52 & 44.66  \\
        \qquad $\triangleright$ \color{midgray}{FedProx} & 77.47 & 67.00 & 14.87 & 22.12 & 41.00 &  44.33 \\
        \qquad $\triangleright$ \color{midgray}{FedAdam} &  77.00 & 67.89 & 15.12 & 22.44 & 42.42 & 44.77\\
        \qquad $\triangleright$ \color{midgray}{FedDyn}  & 78.24 & 69.00 & 16.21 & 21.65 & 42.02 &  45.02\\
        \qquad $\triangleright$ \color{midgray}{FedPer}  & \secondres{79.23} & \secondres{69.20} & \secondres{16.77} & \boldres{22.54} & \boldres{43.43} &  \secondres{46.03}\\
        \qquad \textcolor{red}{$\blacktriangleright$} \color{midgray}{\textbf{\texttt{OmniFM}}} & \boldres{82.33} & \boldres{70.21} & \boldres{18.24} & \secondres{22.35} & \secondres{43.37} &  \boldres{47.30}\\
        
        \textbf{Adapter}~\cite{houlsby2019parameter}  \\
        \qquad $\triangleright$ \color{midgray}{FedAvg} & 72.82 & 64.45 & 11.56 & 21.00 & 38.35 & 41.64 \\
        \qquad $\triangleright$ \color{midgray}{FedProx} & 73.10 & 64.70 & 11.72 & 21.18 & 38.60 & 41.86\\
        \qquad $\triangleright$ \color{midgray}{FedAdam} & 73.05 &  65.12 & 12.08 & 21.47 & 39.12 & 42.17\\
        \qquad $\triangleright$ \color{midgray}{FedDyn}  &  74.02 & 65.88 & 12.41 & 21.39 & 39.05 & 42.55 \\
        \qquad $\triangleright$ \color{midgray}{FedPer}  & \secondres{75.10}  & \secondres{66.30}  & \secondres{12.95}  & \secondres{21.76}  & \secondres{39.70}  & \secondres{43.16} \\
        \qquad \textcolor{red}{$\blacktriangleright$} \color{midgray}{\textbf{\texttt{OmniFM}}} & \boldres{77.85} & \boldres{67.42} & \boldres{13.79} & \boldres{21.89} & \boldres{40.12} & \boldres{44.21} \\
        \textbf{LayerNorm}~\cite{basu2024strong}  \\
        \qquad $\triangleright$ \color{midgray}{FedAvg} &  69.53 & 62.22 & 10.59 & 22.00 & 36.89 & 40.24 \\
        \qquad $\triangleright$ \color{midgray}{FedProx} & 69.80 & 62.40 & 10.75 & 22.12 & 37.05 & 40.42 \\
        \qquad $\triangleright$ \color{midgray}{FedAdam} & 70.02 & 62.95 & 11.08 & 22.35 & 37.48 & 40.78 \\
        \qquad $\triangleright$ \color{midgray}{FedDyn} & 70.55 & 63.40 & 11.32 & 22.28 & 37.60 & 41.03\\
        \qquad $\triangleright$ \color{midgray}{FedPer} & \secondres{71.22} & \secondres{63.88} & \secondres{11.70} & \secondres{22.55} & \secondres{38.10} & \secondres{41.49} \\
        \qquad \textcolor{red}{$\blacktriangleright$} \color{midgray}{\textbf{\texttt{OmniFM}}} & \boldres{73.10} & \boldres{64.60} & \boldres{12.35} & \boldres{22.61} & \boldres{38.52} & \boldres{42.24} \\
        
        \textbf{LoRA}~\cite{hu2021lora}      \\
        \qquad $\triangleright$ \color{midgray}{FedAvg} & 57.73 & 60.18 & 4.59 & 15.00 & 31.16 & 33.73 \\
        \qquad $\triangleright$ \color{midgray}{FedProx} &  58.10 & 60.05 & 4.80 & 15.21 & 31.45 &33.92 \\
        \qquad $\triangleright$ \color{midgray}{FedAdam} & 58.02 & 60.72 & 5.10 & 15.48 & 31.90 & 34.24 \\
        \qquad $\triangleright$ \color{midgray}{FedDyn}  & 58.30 & 60.70 & 5.05 & 15.10 & 31.60 & 34.15 \\
        \qquad $\triangleright$ \color{midgray}{FedPer}  & \secondres{60.20} & \secondres{61.55} & \secondres{5.80} & \secondres{15.70} & \secondres{32.40} & \secondres{35.13} \\
        \qquad \textcolor{red}{$\blacktriangleright$} \color{midgray}{\textbf{\texttt{OmniFM}}} & \boldres{62.45} & \boldres{62.10} & \boldres{6.40} & \boldres{15.66} & \boldres{32.05} & \boldres{35.73}\\
        
        \textbf{Bias}~\cite{cai2020tinytl}     \\
        \qquad $\triangleright$ \color{midgray}{FedAvg} & 68.25 & 55.61 & 11.17 & 17.00 & 35.47 & 37.50 \\
        \qquad $\triangleright$ \color{midgray}{FedProx} & 68.60 & 55.40 & 11.35 & 17.12 & 35.60 & 37.61 \\
        \qquad $\triangleright$ \color{midgray}{FedAdam} & 68.10 & 56.05 & 11.70 & \secondres{17.35} & \secondres{36.10} & 37.86 \\
        \qquad $\triangleright$ \color{midgray}{FedDyn}  & 68.70 & 56.00 & 11.45 & 17.05 & 35.70 & 37.78 \\
        \qquad $\triangleright$ \color{midgray}{FedPer}  & \secondres{70.10} & \secondres{56.88} & \secondres{12.00} & 17.00 & 36.00 & \secondres{38.40} \\
        \qquad \textcolor{red}{$\blacktriangleright$} \color{midgray}{\textbf{\texttt{OmniFM}}} & \boldres{71.25} & \boldres{57.10} & \boldres{12.35} & \boldres{17.61} & \boldres{36.22} & \boldres{38.91}\\
        \midrule

        FedDAT~\cite{chen2024feddat}    & 72.79 & 63.58 & 12.46 & 23.00 & 38.91 & 42.15 \\   
        PromptFL~\cite{guo2023promptfl}  & 65.63 & 57.56 & 4.78 & 15.00 & 40.26 & 36.65 \\
        F$^3$OCUS~\cite{saha2025f} & \secondres{74.69} & \secondres{60.05} & \secondres{12.84} & \secondres{24.00} & \secondres{42.30} & \secondres{42.78} \\
        \textbf{\texttt{OmniFM} (Ours)} & \boldres{77.85} & \boldres{67.42} & \boldres{13.79} & \boldres{21.89} & \boldres{40.12} & \boldres{44.21} \\
        \bottomrule
    \end{tabular}}
    \vspace{-12pt}
    \label{tab:vqa_task3}
\end{table}

\begin{figure*}[tbh]
    \centering
    \includegraphics[width=0.92\textwidth]{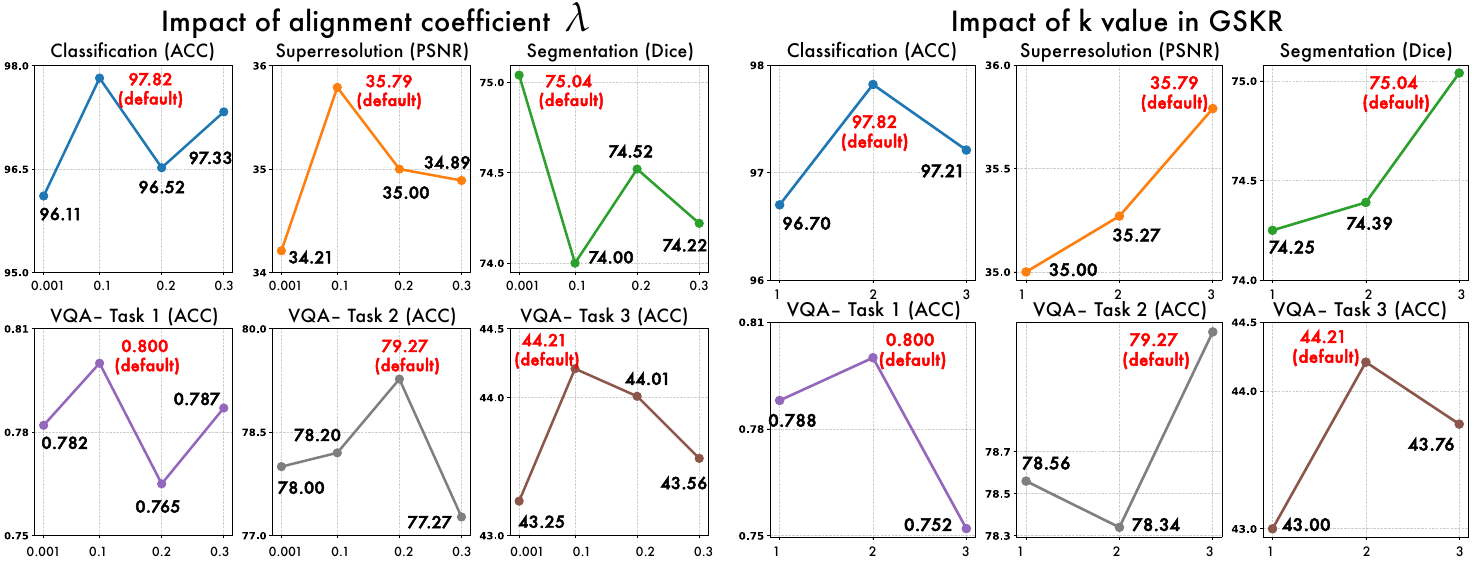}
    \vspace{-7pt}
    \caption{Sensitivity analysis of the alignment coefficient $\lambda$ and the top-$k$ value in GSKR across representative tasks. Default configurations are indicated in red. Due to the paired-image input design, the task of multimodal fusion is evaluated separately.}
    \label{fig:hyper}
    \vspace{-10pt}
\end{figure*}

\subsection{Segmentation}
We evaluate federated segmentation on FeTS2022~\cite{pati2021federated}, a multi-institutional clinical mp-MRI collection for glioma analysis. Following the dataset setup of ~\cite{liu2024fedfms}, we segment GD-enhancing tumor regions from T1ce scans. To simulate intra-modality heterogeneity, we partition clients into four groups with varying institutional compositions (Group 1: clients 2–10; Group 2: 11–15, 17; Group 3: 19–23; Group 4: 19, 22, 23). Each client adopts a standard UNet (default MONAI~\cite{cardoso2022monai} setting), trained for 50 communication rounds with one local epoch per round. The results are shown in Table~\ref{tab:seg}. Across all heterogeneous client groups, our proposed \textbf{\texttt{OmniFM}} achieves the best performance, with especially strong improvements in smaller groups, highlighting superior robustness to distribution imbalance.
\begin{table}[tbh]
  \centering
  \caption{Results on segmentation across 4 intra-modality heterogeneous client groups. \boldres{Bold}: the best, \secondres{Underline}: the second best.}
  \vspace{-8pt}
  \resizebox{.46\textwidth}{!}{
    \begin{tabular}{l|cc|cc|cc|cc|cc}
    \toprule
    \multirow{2}[4]{*}{\bf Method} & \multicolumn{2}{c|}{\makecell{Group 1 \\ (9 Clients)}} & \multicolumn{2}{c|}{\makecell{Group 2 \\ (6 Clients)}} & \multicolumn{2}{c|}{\makecell{Group 3 \\ (5 Clients)}} & \multicolumn{2}{c|}{\makecell{Group 4 \\ (3 Clients)}} & \multicolumn{2}{c}{\makecell{\textbf{Avg.}}} \\
\cmidrule{2-11}          & Dice & IoU & Dice & IoU & Dice & IoU & Dice & IoU & Dice & IoU\\
    \midrule
    Local & 45.24 & 29.33 & 44.21 & 25.00 & 41.25 & 27.22 & 46.23 & 30.00 & 44.23 & 27.89\\
    FedAvg & 44.02 & 28.56 & 44.91 & 25.32 & 40.93 & 26.10  & 46.78 & 30.95 & 44.16 & 27.73\\
    FedProx &  42.32 & 25.23 & 44.00 & 24.43  &  38.99     &   25.32  & 45.35  &  29.76 & 42.66 & 26.19\\
    FedPer & \secondres{71.23} & \secondres{66.99} & \secondres{75.24} & \secondres{62.12} & \secondres{72.43} & \secondres{57.34} & \secondres{76.54} & \secondres{62.99} & \secondres{73.86} & \secondres{62.36}\\
    \midrule
    \textbf{\texttt{OmniFM}} & \boldres{79.84} & \boldres{74.93}  & \boldres{80.82} & \boldres{68.74} & \boldres{78.62} & \boldres{66.01} & \boldres{78.03} & \boldres{65.31} & \boldres{77.41} & \boldres{68.75}\\
    \bottomrule
    \end{tabular}}
  \label{tab:seg}%
  \vspace{-12pt}
\end{table}

\subsection{Multi-modal Fusion}
The multimodal fusion dataset follows~\cite{xu2024simultaneous} and is derived from the Harvard Medical School brain-disease imaging collection, comprising MRI, PET, CT, and SPECT modalities at a resolution of 256. For the inter-modality heterogeneous setup, Client 1 performs CT–MRI fusion, Client 2 performs PET–MRI fusion, and Client 3 performs SPECT–MRI fusion. We adopt U2Fusion~\cite{xu2020u2fusion} as the local backbone and train for 600 rounds with 15 local epochs per round. As shown in Table~\ref{tab:fusion}, \textbf{\texttt{OmniFM}} achieves the optimal performance scores, demonstrating superior preservation of both structural integrity and perceptual fidelity. These highlight the benefit of frequency-aware global priors in stabilizing optimization under cross-modality fusion heterogeneity.

\begin{table}[tbh]
    \centering
    \caption{Results on multi-modal fusion across heterogeneous client configurations. \boldres{Bold}: the best, \secondres{Underline}: the second best.}
    \vspace{-12pt}
    \resizebox{.465\textwidth}{!}{
    \begin{tabular}{lccccccc}
        \toprule
        \multirow{2}{*}{\textbf{Metric}} & \multirow{2}{*}{\textbf{Client}}
        & {\textbf{FedAvg}} & {\textbf{FedProx}} & {\textbf{Ditto}}
        & {\textbf{FedDyn}} & {\textbf{MOON}} & {\textbf{\texttt{OmniFM}}} \\
        & & ~\cite{mcmahan2017communication} & ~\cite{li2020federated} & ~\cite{li2021ditto} & ~\cite{acar2021federated} & ~\cite{li2021model} & \textbf{Ours} \\
        \midrule
        
        \multirow{4}{*}{\textbf{VIF} $\uparrow$}
        & CT-MRI & \boldres{0.221} & 0.185 & 0.196 & 0.182 & 0.195 & \secondres{0.217} \\
        & PET-MRI & \boldres{0.331} & 0.242 & 0.275 & 0.239 & 0.275 & \secondres{0.309} \\
        & SPECT-MRI & \boldres{0.284} & 0.235 & 0.249 & 0.245 & 0.249 & \secondres{0.280} \\
        \cmidrule(lr){2-8}
        & Avg & \boldres{0.279} & 0.221 & 0.240 & 0.222 & 0.240 & \secondres{0.269} \\
        \midrule
        
        \multirow{4}{*}{\textbf{SSIM} $\uparrow$}
        & CT-MRI & \boldres{0.739} & 0.684 & \secondres{0.718} & 0.716 & 0.718 & \boldres{0.739} \\
        & PET-MRI & \secondres{0.640} & 0.550 & 0.609 & 0.617 & 0.609 & \boldres{0.648} \\
        & SPECT-MRI & \boldres{0.749} & 0.702 & 0.732 & \secondres{0.733} & 0.732 & \boldres{0.749} \\
        \cmidrule(lr){2-8}
        & Avg & \secondres{0.709} & 0.645 & 0.686 & 0.689 & 0.686 & \boldres{0.712}\\
        \midrule
        
        \multirow{4}{*}{\textbf{L1} $\downarrow$}
        & CT-MRI & \secondres{0.163} & 0.173 & 0.165 & 0.169 & 0.165 & \boldres{0.161} \\
        & PET-MRI & \secondres{0.189} & 0.199 & \secondres{0.189} & 0.195 & 0.190 & \boldres{0.187} \\
        & SPECT-MRI & \secondres{0.088} & 0.096 & 0.089 & 0.092 & 0.089 & \boldres{0.086} \\
        \cmidrule(lr){2-8}
        & Avg & \secondres{0.147 }& 0.156 & 0.148 & 0.152 & 0.148 & \boldres{0.145} \\
        \bottomrule
    \end{tabular}}
    \label{tab:fusion}
    \vspace{-8pt}
\end{table}

\subsection{Framework Analysis}

\noindent \textbf{Hyperparameter Sensitivity.} The hyperparameter sensitivity  results is shown in Fig.~\ref{fig:hyper}. Across all tasks, performance peaks near the default alignment coefficient, while excessively small or large values introduce under- or over-regularization, respectively. Similarly, $k=2$ yields the most effective global spectral retrieval, as overly small $k$ limits prototype diversity and larger $k$ introduces noisy spectral cues. These indicate that \textbf{\texttt{OmniFM}} remains robust to moderate hyperparameter variation, with clear optima that balance modality alignment and representation fidelity under intra-modality and cross-modality heterogeneity.

\noindent \textbf{Ablation Study.} The results are shown in Table~\ref{tab:ablation}. For fairness, we remove each component while preserving end-to-end functionality: GSKR is replaced with a mean spectral embedding, EWCA with an FiLM layer~\cite{perez2018film}, and PSP with a projection layer to maintain token dimensionality. The training protocol remain unchanged. Removing any module leads to consistent performance drops across tasks, demonstrating that each component contributes to robust cross-modality alignment and generalization.

\begin{table}[tbh]
\centering
\caption{Ablation on representative tasks. 
CLS: Cross-modality classification; SR: Super-resolution; SEG: Segmentation.}
\vspace{-10pt}
\label{tab:ablation}
\resizebox{0.48\textwidth}{!}{
\begin{tabular}{lccccccc}
\toprule
Variant & CLS & SR & SEG & VQA-T1 & VQA-T2 & VQA-T3  \\
\midrule
\textbf{\texttt{OmniFM}} & \textbf{97.82} & \textbf{35.79} & \textbf{75.04} & \textbf{0.800} & \textbf{79.27} & \textbf{44.21}  \\
\midrule
\quad \emph{w/o} GSKR     & 95.82 & 34.88 & 73.42 & 0.775 & 78.91 & 42.62  \\
\quad \emph{w/o} ECA     & 96.51 & 35.04 & 74.03 & 0.783 & 79.15 & 43.01 \\
\quad\emph{w/o} PSP      & 96.92 & 35.12 & 74.51 & 0.787 & 79.42 & 43.19 \\
\quad \emph{w/o} SPAlign  & 97.14 & 35.21 & 74.09 & 0.781 & 77.64 & 42.37 \\
\bottomrule
\end{tabular}}
\vspace{-10pt}
\end{table}

\section{Conclusion}
This paper proposed \textbf{\texttt{OmniFM}}, a unified frequency-aware federated framework for heterogeneous medical imaging. By leveraging global spectral retrieval and spectral-proximal alignment, it effectively mitigates modality-induced divergence and stabilizes optimization. Experiments demonstrate consistent gains across diverse modalities and tasks, confirming its robustness and scalability.


{
    \small
    \bibliographystyle{ieeenat_fullname}
    \bibliography{main}
}


\end{document}